\RequirePackage{fix-cm}
\documentclass[twocolumn]{svjour3}          
\smartqed  
\usepackage{graphicx}
\usepackage{amsmath}
\usepackage{caption}
\usepackage{subcaption}
\usepackage{amssymb}
\usepackage{multirow}
\usepackage[table]{xcolor}
\usepackage{times}
\usepackage{adjustbox}
\usepackage{textcomp}
\usepackage{hhline}
\usepackage{bbding}
\usepackage{cite}
\usepackage{algorithm}
\usepackage{algpseudocode}
\usepackage{enumitem}
\usepackage{wrapfig}
\usepackage{biography}
\newcommand{\specialcell}[2][c]{%
  \begin{tabular}[#1]{@{}c@{}}#2\end{tabular}}
 
\algnewcommand\algorithmicinput{\textbf{Input:}}
\algnewcommand\INPUT{\item[\algorithmicinput]}
 
\begin{document}

\title{Microwave breast cancer detection using Empirical Mode Decomposition features
}



\author{Hongchao Song$^{1}$         \and
        Yunpeng Li$^{2}$ \and Mark Coates$^{2}$ \and Aidong Men$^{1}$ 
}

\institute{\Envelope Hongchao Song \and Aidong Men \at
$^{1}$School of Information and Communication Engineering, Beijing University of Posts and Telecommunications, Beijing, China\\
              \email{shch@bupt.edu.cn, menad@bupt.edu.cn}           
           \and
          Yunpeng Li \and Mark Coates \at
              ${^2}$Dept. of Electrical and Computer Engineering, McGill University, Montr\'eal, Qu\'ebec, Canada\\
              \email{yunpeng.li@mail.mcgill.ca, mark.coates@mcgill.ca}
}

\date{Received: date / Accepted: date}

\maketitle

\begin{abstract}
 Microwave-based breast cancer detection has been proposed as a
  complementary approach to compensate for some drawbacks of
  existing breast cancer detection techniques.  Among the existing
  microwave breast cancer detection methods, machine learning-type
  algorithms have recently become more popular. These focus on detecting
  the existence of breast tumours rather than performing imaging to
  identify the exact tumour position. A key step of the machine learning
  approaches is feature extraction. One of the most widely used
  feature extraction method is principle component analysis
  (PCA). However, it can be sensitive to signal misalignment. This
  paper presents an empirical mode decomposition (EMD)-based feature
  extraction method, which is more robust to the
  misalignment. Experimental results involving clinical data sets
  combined with numerically simulated tumour responses show
  that combined features from EMD and PCA improve the detection
  performance with an ensemble selection-based classifier.
\end{abstract}

%
%
%
%
\section{Introduction}
\label{intro}

Detecting breast cancer during the early stages of development is one
of the best ways to improve the recovery rate~\cite{CCS2016}. Among
the widely used breast cancer screening methods, magnetic
resonance imaging (MRI) is very expensive, and
X-ray mammography involves exposure to radiation and painful breast
compression. Neither can be frequently used as a routine detection method. Ultrasound is
inexpensive but is less accurate and can struggle to differentiate
between benign and malignant lesions~\cite{lim2008}. Microwave-based breast cancer detection
methods have been intensively explored over the past two decades as a
potential complementary screening modality. The basis of this method
is the reported significant differences in dielectric properties between malignant
 and healthy breast tissue in the microwave frequency range
\cite{hagness2000,sugitani2014}.

In the existing research in the field of microwave breast cancer
screening, two different directions have been explored: imaging
methods
\cite{caorsi1993,meaney2007,shea2010,lim2008,byrne2015,halloran2010,bourqui2012,klemm2010,li2001}
and machine learning methods
\cite{alshehri2011,conceical2013,concei2014,li2015,byrne2011,davis2008}. In
both cases, transmitting antennas illuminate the breast area using
ultra-wideband microwave signals and receiving antennas record the
reflected signals. In the imaging methods, algorithms are applied to
construct an image of the breast that can be used to determine the
existence and position of a tumour. The imaging algorithms fall
into two categories: tomography and radar (backscatter). The
tomography methods try to estimate the electromagnetic parameters of
the entire breast using inverse scattering algorithms. This is an
ill-posed problem, and leads to challenges in robust inference and a
high computational cost \cite{caorsi1993,meaney2007,shea2010}. Radar
imaging methods
\cite{lim2008,byrne2015,halloran2010,bourqui2012,klemm2010,li2001} use
confocal algorithms to determine the position of the significant
scatterers in the breast. These indicate regions of high contrast in
dielectric properties that may be caused by the presence of malignant
tissue. Alternative imaging algorithms, based respectively on Bayesian
modeling and time reversal techniques, have been proposed in
\cite{davis2005,kosmas2006}.

Another direction is the application of machine learning
techniques. Here the goal is usually detection of the presence of a
tumour, as opposed to identifying its location. A key component of machine learning algorithms is feature
extraction. This process generates useful features in a lower
dimension than that of the original data. In a wide variety of
applications, the quality of the extracted features has been shown to
have a dramatic effect on the performance of classifiers.
Thus, the feature extraction method adopted should be robust to
the data characteristics in the target application.

Local discriminant bases (LDB) were used to identify features
in~\cite{davis2008}; these strive to find orthonormal principal
components of the data with localized time-frequency
characteristics. The discrete cosine transform (DCT) was applied to
extract features in \cite{alshehri2011}.  The most widely used feature
extraction method in microwave breast cancer detection is principal
component analysis
(PCA)~\cite{conceical2013,concei2014,li2015,byrne2011,davis2008}.  PCA
reduces the data dimension by projecting the data onto a subspace that
retains the maximum variance of the signal collection. The PCA methods
have provided the most promising results, but a significant deficiency
is the sensitivity to misalignment between signals.  Misalignment is
common in microwave breast scans due to mechanical issues such as
antenna movement between scans and the system intrinsic jitter in the
clock and the sampling oscilloscope
\cite{santorelli2013,porter2013B}. Thus, there can be a great
disparity between the features extracted from two scans of the same
healthy breast, arising simply because the recordings are slightly
shifted in time. In most of the previously reported
results~\cite{conceical2013,concei2014,byrne2011,davis2008}, this has
not been a significant issue because the analysis used numerically
simulated data or phantom measurements. 

Inspired by the successful application of empirical mode decomposition
(EMD) in Electroencephalography (EEG) signal processing
\cite{bajaj2012,li2013}, we explore the incorporation of
EMD into our feature extraction method for microwave breast cancer
detection. The EMD method is a data-driven technique that
decomposes the non-stationary signal into a 
usually small number of intrinsic mode functions (IMFs)~\cite{huang1998}.
The EMD method is insensitive to misalignment of signals,
as it is applied to each signal independently.
If we shift a signal by a few time samples, the IMFs
obtained from the shifted signal will be time-shifted copies of those obtained
from the original signal. If we specify features as statistical
properties of the IMFs that are insensitive to the time-shift, such as
mean absolute values and standard deviations, then the feature extraction procedure
is robust to the presence of jitter in the measurement process.
In contrast, PCA processes multiple signals jointly;
changing any signal results in different PCA components and scores for
all measurements.

Given these desired characteristics of EMD, we propose combining
EMD-based features with PCA-based features for microwave breast tumour
detection, to exploit the complementary strengths of both feature
extraction approaches.  EMD-based feature extraction focuses on the
characteristics of a single measurement. PCA-based feature extraction
focuses on the variances among different samples. We test the
performance of the proposed method and discuss the results obtained in
our experiments.

\section{Methods}
\label{sec:materials}

\subsection{Feature extraction and classification}
\label{sec:features extraction}

\subsubsection{Empirical mode decomposition}

The empirical mode decomposition is an
iterative procedure, as described in
Algorithm~\ref{alg:EMD}~\cite{huang1998}.
During each iteration a decomposition signal is
constructed. This decomposition signal is identified as an IMF if it
satisfies two conditions:
\begin{description}
\item [{\em C1}:] The difference between the number of extrema (maxima and minima) and the number of zero-crossings must be no more than one;
\item [{\em C2}:] The local mean, defined as the mean of the upper and lower envelopes, must be zero. 
\end{description}

\begin{algorithm}[tbh]
\begin{algorithmic}[1]
\INPUT Signal $x$. 
\INPUT Maximum number of IMFs: $N_{\mathrm{IMF}}$.
\State Set $k=0$ and $r_0=x$. Set $z=x$.
\While {$k<N_{\mathrm{IMF}}$}
\State Find all extrema of $z$.
\State Interpolate between minima (maxima) of $z$ to obtain the
lower (upper) envelope $e_{\min}$ ($e_{\max}$).
\State Compute the mean envelope $m=(e_{\min}+e_{\max})/2$.
\State Compute the IMF candidate $d_{k+1}=z-m$.
\State Is $d_{k+1}$ an IMF ? (does it satisfy {\em C1} and {\em C2}?)
\begin{itemize}
    \item Yes. Save $d_{k+1}$; compute the residue
      $r_{k+1}=x-\sum_{i=1}^kd_i$. Set $z=r_{k+1}$.
    \item No. Set $z = d_{k+1}$.
\end{itemize}
\EndWhile
\end{algorithmic}
\caption{Empirical mode decomposition}
\label{alg:EMD}
\end{algorithm}

\begin{figure}[htbp]
\centering
\includegraphics[width=\linewidth]{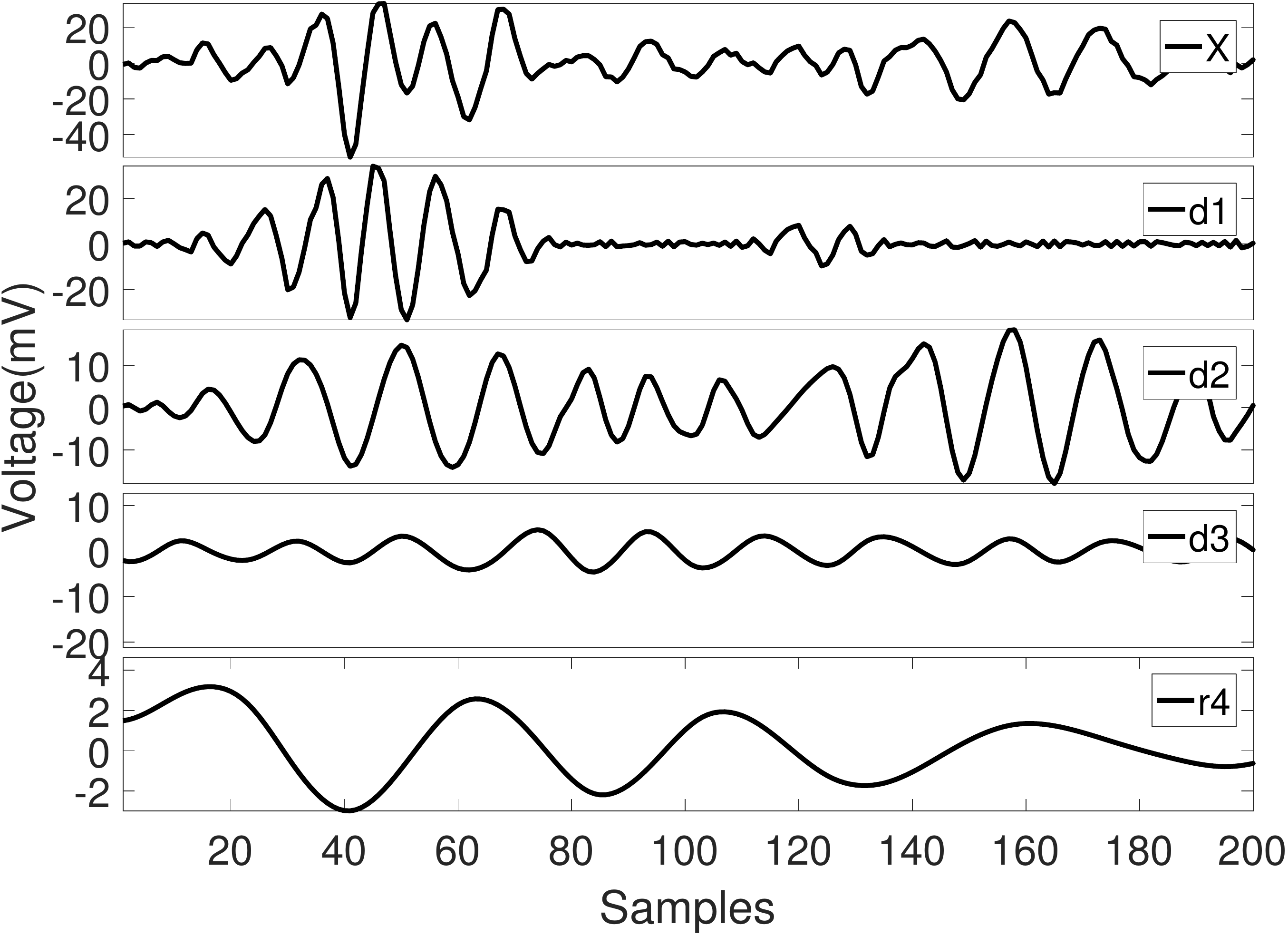}
\caption{Example of EMD. X is a healthy measurement collected by the
  antenna pair A16A15 from the left breast of volunteer two during her
  first visit; $d1-d3$ are the first three IMFs (modes); and $r4$ is the residue.}
\label{fig:EMD}
\end{figure}

An example of the EMD decomposition of an antenna-pair recording is
shown in Figure \ref{fig:EMD}.  The IMFs can be viewed as a set of
components of the signal, each capturing different time-scale
characteristics and local structure.~\cite{huang1998}.
\subsubsection{Feature extraction}

Following the practice in~\cite{li2013}, we calculate the first 5 IMFs
of the EMD for each measurement. For each IMF, 4 features are
calculated. They are the mean absolute value ($\mu$), the standard deviation
($\sigma$), the kurtosis ($\kappa$) and the mean absolute successive difference ($S$). The mean absolute successive difference is used to measure the
intensity of signal changes. The features for the $k$-th IMF, of length $L$
time samples, are:

\begin{enumerate}
\item Mean absolute value ($\mu_k$):
\begin{equation}
\mu_k=\frac{1}{L}\sum_{i=1}^L\left| {d_k(i)}\right|\,,
\end{equation}

\item Standard deviation ($\sigma_k$):
\begin{equation}
\begin{aligned}
\sigma_k=\sqrt{\frac{1}{L}\sum_{i=1}^L( {d_k(i)}-\frac{1}{L}\sum^L_{i}d_k(i))^2}\,,
\end{aligned}
\end{equation}

\item Kurtosis ($\kappa_k$):
\begin{equation}
\begin{aligned}
M_k=\frac{1}{L}\sum^L_{i=1}d_k(i)\\
\kappa_k=\frac{\frac{1}{L}\sum^L_{i=1}(d_k(i)-M_k)^4}{
(\frac{1}{L}\sum^L_{i=1}(d_k(i)-M_k)^2)^2}
\end{aligned}
\end{equation}

\item Mean absolute successive difference ($S_k$):
\begin{equation}
S_k=\frac{1}{L-1}\sum_{i=1}^{L-1}\left|{d_k(i+1)}-d_k(i)\right| \,.
\end{equation}

\end{enumerate}

The magnitudes of the features derived from IMFs obtained later in the
decomposition procedure are most often smaller in magnitude. For the measurements
which are collected by the same antenna pair, we rescale each feature
of the training data into the range $[0,1]$. The testing data is
scaled by the same ratio. The process is repeated for each antenna
pair.

We also perform PCA on the signals derived from each antenna pair. 30
principal scores are retained, as in~\cite{li2016}. The PCA scores are
also normalized, by scaling the first principal components to the
range $[0,1]$. The PCA scores of the other dimensions are rescaled by
the same ratio. For PCA features, a coefficient matrix is calculated
from the training data. This is formed from the measurements collected
by the same antenna pair from all volunteers, and both healthy
measurements and tumour response-injected measurements are
included. The PCA features associated with a single antenna pair
measurement are thus dependent on all of the other measurements for
that antenna pair. In contrast, the EMD features corresponding to a
single time-series recorded by an antenna pair are independent of any
other measurements. When we consider combining the features, each
antenna pair time series measurement is represented by a 50
dimensional feature vector, combining the 30 PCA features and the 20
EMD features.

\subsubsection{Classification}

\begin{figure}
\centering
\includegraphics[width=\linewidth]{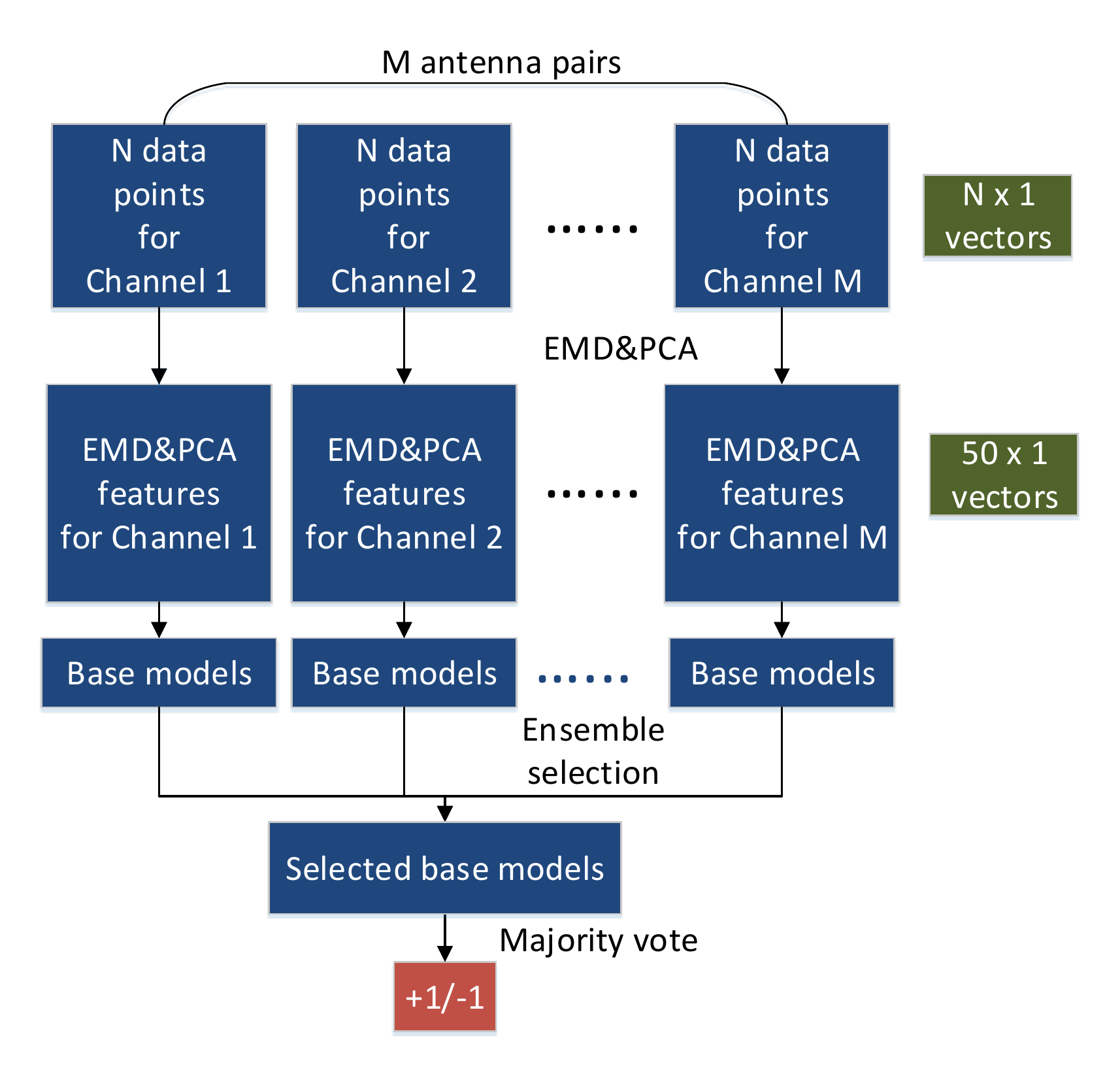}
\caption{The combined feature extraction and classification procedure.}
\label{fig:whole procedures}
\end{figure}

The classifier we adopt is the cost-sensitive ensemble selection-based
classifier proposed in~\cite{li2016}. This demonstrates better
classification performance than imaging-based classifiers and
other ensemble classification structures. The ensemble selection-based
classifier constructs base classifiers by choosing different antenna
pairs and different hyper parameters in a cost-sensitive $2\nu$-
support vector machine (SVM) \cite{chew2001}. The hyper-parameters
$\nu_+$ and $\nu_-$ control the importance associated with the two
types of classification error (false-positive and false-negative)
while the hyper-parameter $\gamma$ is the kernel width of the SVM.
The model selection stage chooses the base models with the smallest
Neyman-Pearson measure $\hat{e}$~\cite{scott2007}
\begin{equation}
\hat{e}=\frac{1}{\alpha}\max\{\hat{P}_F-\alpha,0\}+\hat{P}_M\,,
\label{fun:NP}
\end{equation}
where $\hat{P}_M$ is the empirical miss probability, $\hat{P}_F$ is
the empirical false positive rate, and $\alpha$ is the upper bound of
the target false positive rate. The classification
algorithm is depicted in Figure \ref{fig:whole procedures}.

\subsection{Experimental data}
\label{sec:data}

The experimental data were collected with a multi-static radar system
with $R=16$ antennas~\cite{porter2013A} built by the research team at
McGill University.  For one scan, each of the $R$ antennas transmits
an ultra-wideband pulse into the breast, and the other antennas record
the backscattered signals. This process is repeated until one
measurement is recorded by each of the $M=R(R-1)=240$ different
antenna pairs.

We performed breast scans on 12 healthy volunteers. Table
\ref{table:nVisits} lists the number of visits for each volunteer. The
clinical trial lasted 8 months, and involved 48 patient visits, with
each volunteer visiting at most once per month~\cite{porter2016}. We
collected measurements of the left breast and the right breast from
the same person during each visit.  Thus, $48\times 2=96$ sets of
measurements were collected.
\begin{table}[htbp]
\centering
\caption{\textnormal{Number of visits for each volunteer}}
\begin{tabular*}{\linewidth}{@{\extracolsep{\fill}}ccccccc}
\hline
Volunteer index& 1&2 &3 &4 &5 &6\\
Number of visits&3 &3 &4 &5 &2 &6\\
\hline
Volunteer index&7&8 &9 &10 &11 &12\\
Number of visits&6 &4 &4 &4 &3 &4\\
\hline
\end{tabular*}
\label{table:nVisits}
\end{table}

Since all participants of the clinical trial were healthy volunteers,
there are only tumour-free measurements in our original data set.  We
adopt the strategy outlined in~\cite{li2016} to simulate tumour responses for
each volunteer, based on the transmitted pulses from the antennas and
the dielectric properties of breast tissue. For one antenna pair and a
tumour position $p_0$, the frequency domain representation of the
tumour response $R^t(p_0,\omega)$ is modeled as:
\begin{equation}
\label{equation:tumour_response}
R^t(p_0,\omega)=\Gamma R(\omega)e^{-j(k_{im}(d^t_{im}-d^d_{im})+k_{br}(d^t_{br}-d^d_{br}))},
\end{equation}
where $R(\omega)$ is the frequency domain representation of received
signal, $d^d_{im}$ and $d^d_{br}$ are the lengths of the direct path
for this antenna pair through the immersion medium (ultrasound gel)
and breast tissue, respectively. $d^t_{im}$ and $d^t_{br}$ are the
lengths of the shortest path between the antenna pair via the
tumour position $p_0$, in the immersion medium and the breast tissue,
respectively. $\Gamma$ is a constant that can be used to introduce
additional attenuation in the tumour response. In this paper, we
concentrate on the case $\Gamma = 1$.  $k_{im}$ and $k_{br}$ are the
wavenumbers for the immersion medium and breast tissue, respectively,
and these have the following expressions:
\begin{align}
k_{im}(\omega) &= \frac{2\pi}{\lambda_{im}(\omega)} =\sqrt{\epsilon_{im}(\omega)}\frac{\omega}{c}\,,
\label{equation:kim} \\
k_{br}(\omega) &=\frac{2\pi}{\lambda_{br}(\omega)} =\sqrt{\epsilon_{br}(\omega)}\frac{\omega}{c}\,.
\label{equation:kbr}
\end{align}
Here $\epsilon_{im}$ is the relative permittivity of the immersion
medium, $c$ is the speed of light, and $\epsilon_{br}$ is the average breast tissue relative
permittivity. The latter is specified by the Debye model \cite{debye1929}:
\begin{equation}
\epsilon_{br}(\omega)=\epsilon_{\infty}+\frac{\Delta\epsilon}{1+j\omega\tau}+\frac{\sigma_s}{j\omega\epsilon_0}\,,
\label{equation:Debye}
\end{equation}
where $\epsilon_0=8.854\times10^{-12}$ F/m is the permittivity of free
space, and $\epsilon_{\infty}$ is the dielectric constant of the
material at infinite
frequency. $\Delta\epsilon=\epsilon_{s}-\epsilon_{\infty}$, and
$\epsilon_s$ is the static dielectric constant. The pole relaxation
constant is $\tau$ and the static conductivity is
$\sigma_{s}$. The model parameters are chosen to approximate the
dielectric properties of breast tissue.

\begin{figure}[htbp]
\centering
\includegraphics[width=\linewidth]{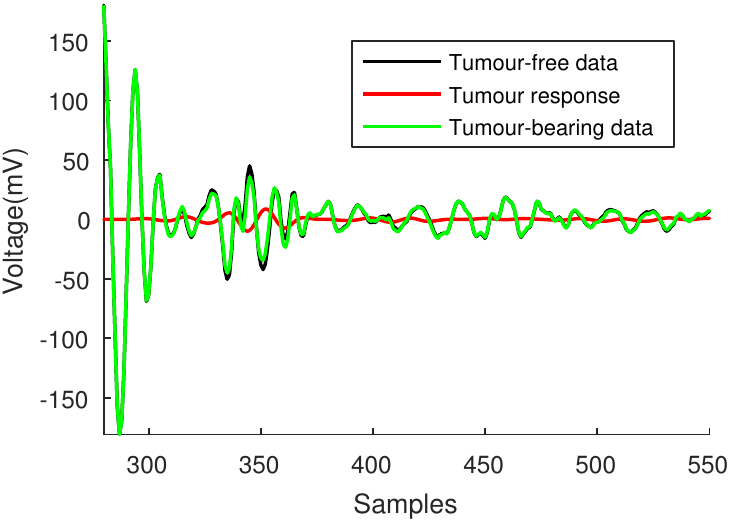}
\caption{An example of received pulses before and after tumour
  response injection. The pulse is collected by the antenna pair A1A2 from the left breast of volunteer one during her first visit.}
\label{fig:tumour_injection}
\end{figure}

We refer readers to~\cite{li2016} for a more complete description and
discussion of the tumour response simulation procedure. We do stress
that this is not a model that assumes homogeneous breast tissue. By
including the {\em received} signal $R(w)$, rather than the
transmitted signal, the model incorporates distortions and delays
caused by inhomogeneous tissue.  In Figure \ref{fig:tumour_injection}, we
present one example of a tumour response injected signal.

In order to evaluate the classification performance with
different feature extraction methods, we generate $50$ datasets with
different breast relative permittivity values and different tumour
positions for the same volunteer.  For each data set, half of each
volunteer's visits are randomly selected as tumour-bearing visits. If
the number of visits $N_j$ for patient $j$ is an odd number, $\frac{N_j-1}{2}$ visits are
selected as tumour-bearing visits. In these tumour-bearing visits, we
randomly select one breast as the tumour-bearing breast, and randomly
sample a tumour position in the upper-quadrant region of the breast,
closer to the armpit, which is reported to be the most likely breast tumour region
\cite{sohn2008}.  The Debye model parameters are randomly generated
for different volunteers. The ranges of possible parameter values are
detailed in~\cite{li2016}. The breast permittivities are different for
different volunteers.

\begin{figure}[htbp]
\centering
\includegraphics[width=\linewidth]{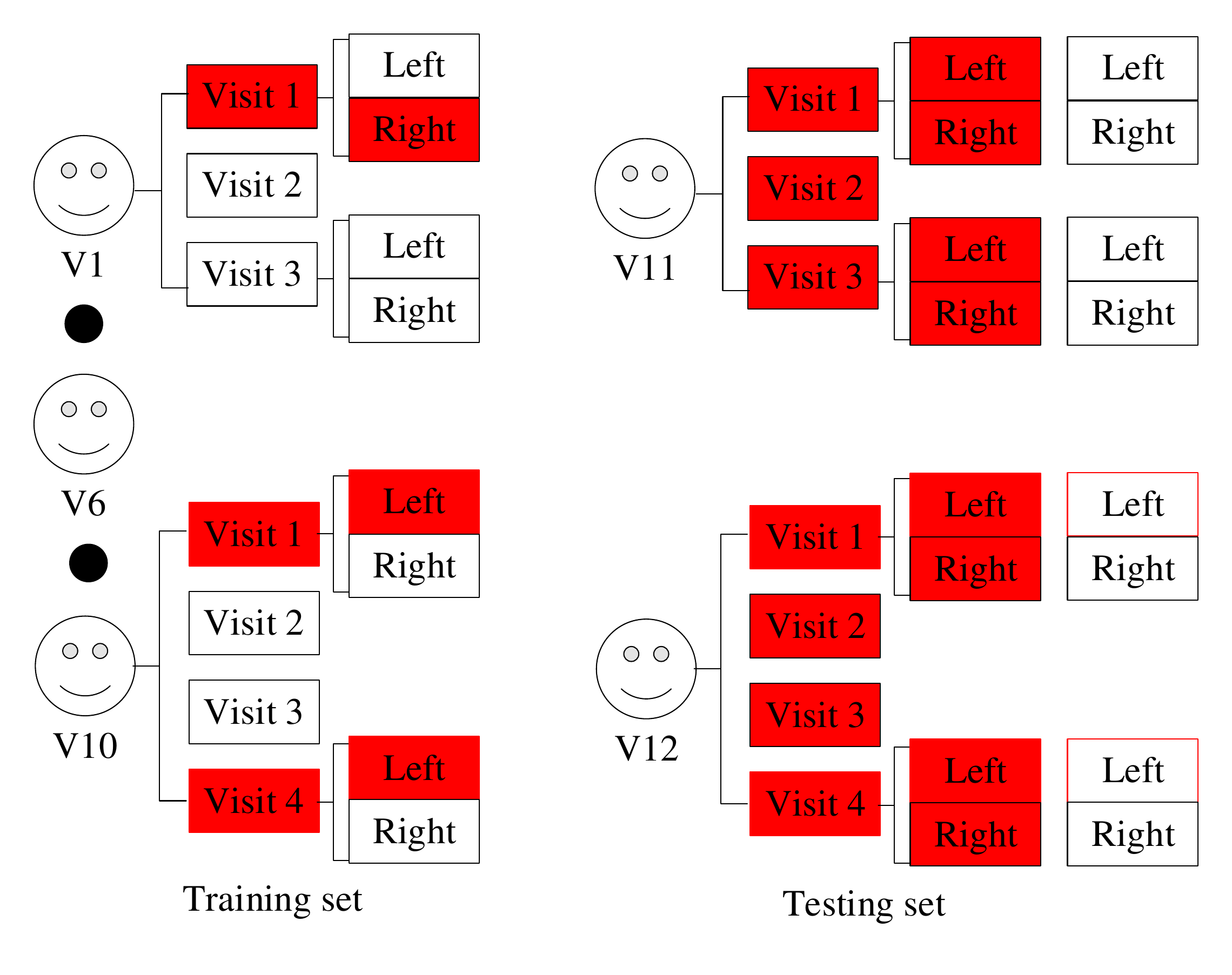}
\caption{One training-testing pair in which red blocks indicate randomly selected tumour visits and simulated tumour-bearing breasts, and white blocks indicate healthy visits and healthy breasts.}
\label{fig:dataset}
\end{figure}

In evaluating the classification performance, we divide each
data set into $\binom{12}{2}=66$ training-testing pairs,
by using data from 10 volunteers to form the training data,
and data from 2 volunteers to form the testing data.
To construct testing data, we further add tumour-bearing scans for
each tumour-free scan, so that the ratio of tumour-free
scans and tumour-bearing scans is 1:1 in the test data. This allows us
to more extensively test classification performance.
To achieve this, we randomly generate tumour positions for each scan
so that different tumour positions may exist for the same volunteers.
However, we keep the same relative permittivity for the same volunteer
in the training and testing set.
We do not add more tumour-bearing data in the training data, so
that the ratio of tumour-free
scans and tumour-bearing scans in the training data is 3:1.
This forms an imbalanced training data set to train the classifier
and mimics the real scenario where the number of tumour-free
scans is more than the number of tumour-bearing scans.
A graphical illustration for one training-testing pair of one data set
is provided in Figure~\ref{fig:dataset}. 

\subsection{Parameter settings}

As discussed in~\cite{li2016}, signals collected by the antenna
pairs located on the opposite sides of the breast can be highly
distorted and vary significantly among different volunteer
visits. Thus, we discard signals from any antenna pair whose
median peak amplitude of all the training data is less than a
threshold of $20$ mV. 
The signals are also truncated for the detection algorithm to focus on a
region where tumour responses are most likely to occur. The windowing
range is $[61, 600]$ samples. These values are calculated as the
feasible time period in which a tumour response can occur
(see~\cite{li2016}). Excluding values outside this window reduces the
noise during signal processing.

The candidate values of the $2\nu$-SVM hyper-parameters used
for cross validation are listed in Table \ref{table:parameters}.
We tested the detection performance with the $\gamma$ value chosen
from the candidate set $\gamma=\{2^{-15},2^{-13},...,2^{5}\}$ using
a small subset of the data, and observed that the ensemble classifier
almost always chose the $\gamma$ values of $\{2^{-1}, 2^{1}\}$. We
further tested a range of fixed $\gamma$ values between $2^{-2}$
and $2^{2}$ and observed that all gave similar performance. To reduce the computational
cost during training, we set $\gamma= 1$ for the experiments detailed here.

Thus, there are $1 \times 18\times 18=324$ different $2\nu$-SVM hyper-parameter
combinations. These are used to produce a
model library consisting of $M\times 324$ base models, where $M$ is the
number of retained antenna pairs in each data set. The ensemble
classifier selects $100$ base models, choosing those 
with the smallest Neyman-Pearson measure when applied to the training data,
to perform classification on the test data.

\begin{table}[htbp]
\centering
\tabcolsep=0.4cm
\caption{\textnormal{Candidate values of hyper-parameters of $2\nu$-SVM in cross validation.}}
\label{table:parameters}
\begin{tabular}{|c|c|}
	\hline
	Parameter& Value \\ \hline
  \rule{0pt}{10pt}  $\gamma$&1 \\ \hline
    \rule{0pt}{15pt}\multirow{2}{*}{$\nu_{+}$}& \specialcell{$1\times 10^{-5}$,$3\times 10^{-5}$,$1\times 10^{-4}$,$3\times 10^{-4}$,\\$0.001,0.003,0.01,0.03,0.1,0.2,0.3,...,1$} \\ \hline
  \rule{0pt}{15pt} \multirow{2}{*}{$\nu_{-}$}& \specialcell{$1\times 10^{-5}$,$3\times 10^{-5}$,$1\times 10^{-4}$,$3\times 10^{-4}$,\\$0.001,0.003,0.01,0.03,0.1,0.2,0.3,...,1$} \\ \hline
	\end{tabular}
\end{table}

\section{RESULTS}
\label{sec:experiment results}

\begin{figure}[htbp]
\centering
\begin{subfigure}[b]{\linewidth}
\includegraphics[width=\linewidth]{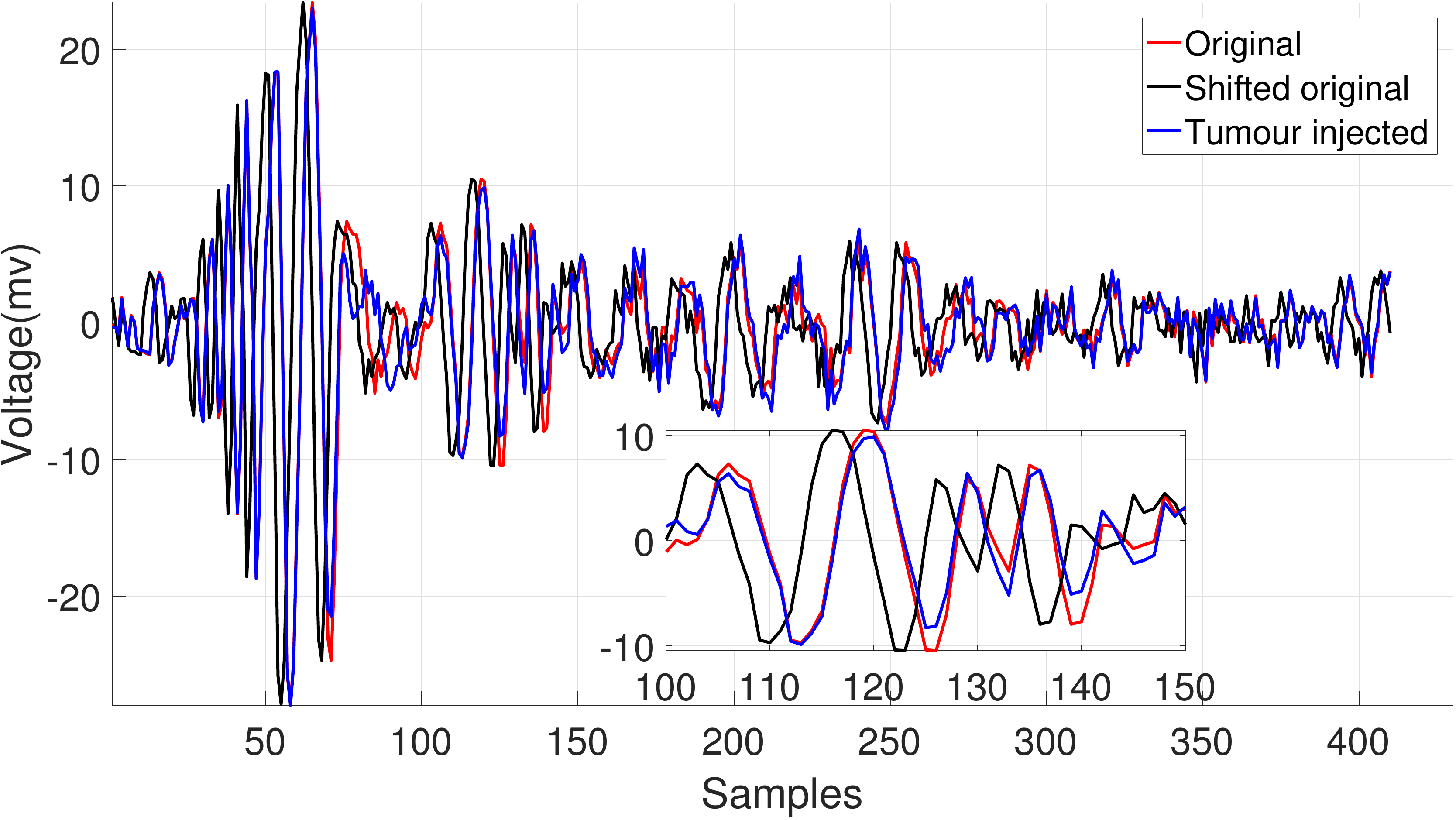}
\centering
\caption{An example of the original measurement, the original
  measurement shifted by four time samples, and the tumour response-injected measurement.}
\label{fig:healthy_tumour}
\end{subfigure}

\begin{subfigure}[b]{\linewidth}
\includegraphics[width=\linewidth]{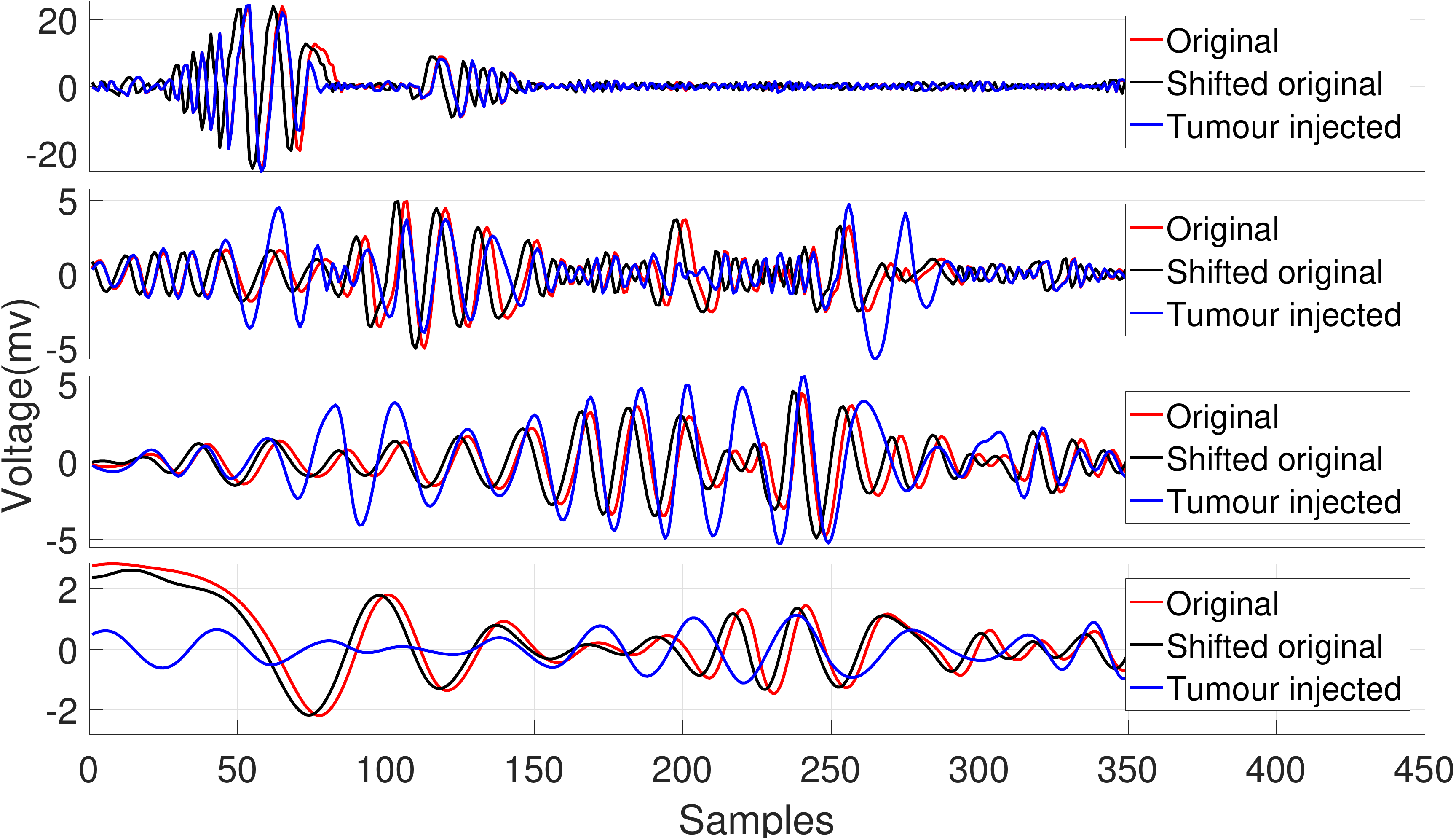}
\centering
\caption{The first four IMFs identified by EMD.}
\label{fig:EMD_2signals}

\end{subfigure}
\caption{Comparisons of IMFs derived from different measurements, including the original measurement, the shifted original measurement and the tumour response-injected measurement.}
\label{fig:EMD insensitive}
\end{figure}

\begin{figure}[htbp]
\centering
\includegraphics[width=\linewidth]{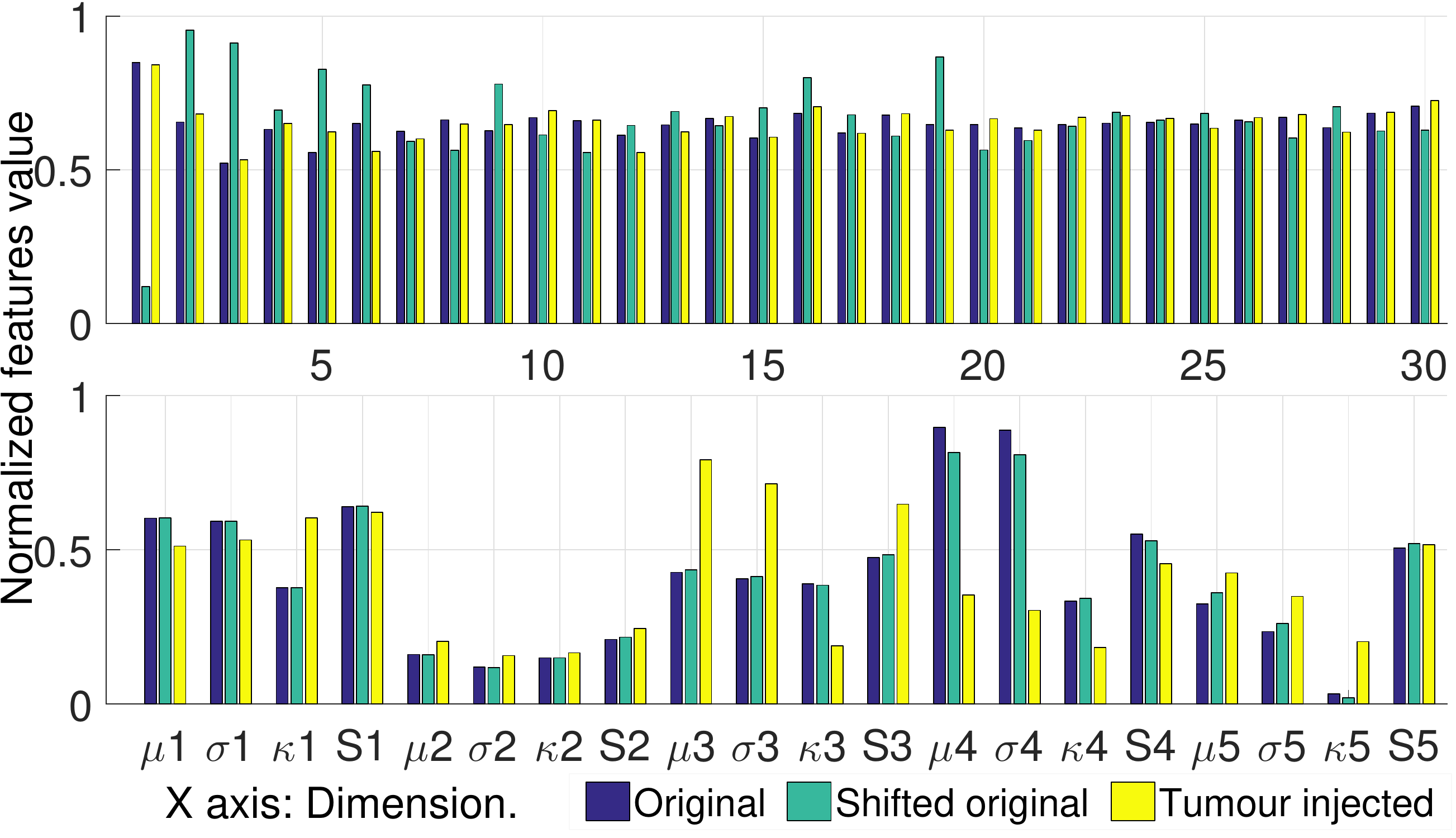}
\caption{Normalized EMD (bottom) and PCA features (top), X axis is dimension index. }
\label{fig:EMD_PCA_features}
\end{figure}

We first verify whether the EMD-based features are
robust to signal misalignment.
The EMD is applied to individual signals, so time shifts in one signal due to misalignment only affect the IMFs of that signal. In addition, the IMFs derived by the EMD are shift-invariant, in the sense that a shift in the input signal, provided it is an integer number of samples, will result in an equivalent shift in the IMFs. The features we have chose to derive from them are insensitive to these shifts.
An example is shown in Figure \ref{fig:EMD insensitive} and Figure \ref{fig:EMD_PCA_features}.

The EMD and PCA features extracted from the measurements in Figure
\ref{fig:healthy_tumour} are shown in Figure
\ref{fig:EMD_PCA_features}.
The figure shows that the PCA features extracted from the original
measurement and those from the tumour response-injected measurement are
similar. The same observation applies for all other antenna pairs
and tumour measurements. The classifier can perform much better if it
is easier to discriminate between the features derived from the
healthy and tumour response-injected measurements. Indeed, for PCA,
there appears to be a greater discrepancy between the features of the
original and time-shifted signals. For the EMD feature extraction, there is less discrepancy between the
features derived from the original and time-shifted signals, and a
greater distinction between the original and tumour response-injected measurements.

We evaluate the performance of the ensemble selection detection
algorithm with EMD extracted features, PCA scores, and combined
feature vectors from EMD and PCA, respectively.  The detection
performance with different $\alpha$ values is shown in Table
\ref{table:FP_FN}. Irrespective of the features used, the ensemble
selection detection algorithm almost always achieves an empirical
false positive rate on the test data set that lies below the target
false positive rate specified by $\alpha$. The use of combined
features (EMD and PCA) leads to the smallest miss probability and also
the smallest Neyman-Pearson error for all three values of $\alpha$.

\begin{table}[htbp]
\centering
\tabcolsep=0.1cm
\caption{\textnormal{Average performance of three different kinds of
features with different target false positive rates $\alpha$.
The $10\%$ and $90\%$ quantiles are shown in square brackets.
The columns show the average false positive rate, the average false negative rate,
the average error, and the average Neyman-Pearson error measure (Equation
\ref{fun:NP}). Shaded entries indicate the smallest Neyman-Pearson error with
the specific $\alpha$}. }
\scriptsize         
\label{table:FP_FN}
\begin{tabular}{|c|c|c|c|c|c|}
	\hline
	$\alpha$ & Method & $\hat{P}_F$ & $\hat{P}_M$ & \mbox{average error} & $\hat{e}$ \\ \hline
    
\rule{0pt}{13pt} \multirow{3}{*}{$\alpha$=0.1} & EMD 
	& $\specialcell{0.03\\$[0.00,0.10]$} $ & \specialcell{0.65\\$[0.44,0.86]$} & \specialcell{0.34\\
	$[0.25,0.44]$}&\specialcell{0.70\\$[0.45,0.93]$}\\ \cline{2-6}
	
\rule{0pt}{13pt} &PCA& $\specialcell{0.01\\$[0.00, 0.07]$}$ & \specialcell{0.63\\$[0.42, 0.83]$}& \specialcell{0.32\\$[0.21,0.42]$} &  \specialcell{0.65\\$[0.43,0.86]$}\\ \cline{2-6}

\rule{0pt}{13pt}	&\specialcell{EMD\\PCA}& $\specialcell{0.02\\$[0.00, 0.10]$}$ & \specialcell{0.59\\$[0.38, 0.81]$}& \specialcell{0.31\\$[0.21,0.42]$} &  \cellcolor{blue!25}\specialcell{0.62\\$[0.39,0.85]$}\\ \hline	
	
\rule{0pt}{13pt} \multirow{3}{*}{$\alpha$=0.3} & EMD 
	& $\specialcell{0.09\\$[0.00,0.22]$} $ & \specialcell{0.43\\$[0.21,0.65]$} & \specialcell{0.27\\
	$[0.16,0.37]$}&\specialcell{0.44\\$[0.22,0.67]$}\\ \cline{2-6}
	
\rule{0pt}{13pt}	&PCA& $\specialcell{0.04\\$[0.00, 0.14]$}$ & \specialcell{0.57\\$[0.33, 0.79]$}& \specialcell{0.30\\$[0.18,0.42]$} &  \specialcell{0.57\\$[0.33,0.78]$}\\ \cline{2-6}

\rule{0pt}{13pt}	&\specialcell{EMD\\PCA}& $\specialcell{0.09\\$[0.00, 0.21]$}$ & \specialcell{0.40\\$[0.18, 0.60]$}& \specialcell{0.24\\$[0.14,0.35]$} &  \cellcolor{blue!25}\specialcell{0.40\\$[0.19,0.62]$}\\ \hline

\rule{0pt}{13pt} \multirow{3}{*}{$\alpha$=0.5} & EMD 
	& $\specialcell{0.21\\$[0.00,0.4]$} $ & \specialcell{0.27\\$[0.08,0.50]$} & \specialcell{0.24\\
	$[0.14,0.35]$}&\specialcell{0.28\\$[0.08,0.5]$}\\ \cline{2-6}
	
\rule{0pt}{13pt}	&PCA& $\specialcell{0.14\\$[0.00, 0.36]$}$ & \specialcell{0.52\\$[0.29, 0.75]$}& \specialcell{0.33\\$[0.19,0.46]$} &  \specialcell{0.53\\$[0.30,0.75]$}\\ \cline{2-6}

\rule{0pt}{13pt}	&\specialcell{EMD\\PCA}& $\specialcell{0.18\\$[0.00, 0.36]$}$ & \specialcell{0.26\\$[0.07, 0.45]$}& \specialcell{0.22\\$[0.10,0.32]$} &  \cellcolor{blue!25}\specialcell{0.27\\$[0.07,0.45]$}\\ \hline
	
\end{tabular}
\end{table}

\begin{figure}[htbp]
\centering
\includegraphics[width=\linewidth]{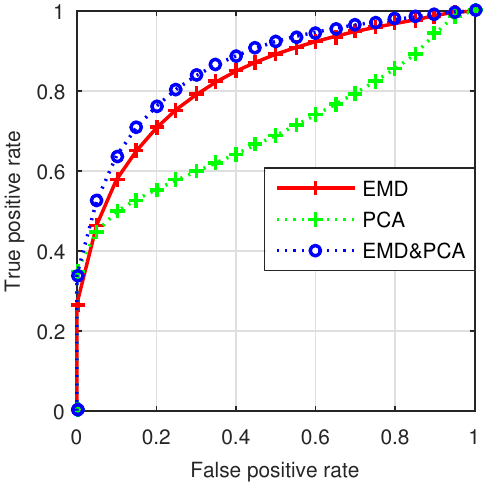}
\caption{Average receiver operating characteristic (ROC) curves for
  different features (EMD, PCA, and combined EMD and PCA)
  for the tumour response-injected clinical data with~$\Gamma=1$. }
\label{fig:ROC}
\end{figure}

If we vary $\alpha$ over $101$ evenly-spaced values between $0$ and
$1$, we obtain different false positive rates and true positive rates
on the test data. These form different operating points in the
receiver operating characteristic (ROC) curves. Figure~\ref{fig:ROC}
shows the ROC curves, averaged over the 50 data sets, for the three
choices of features. The areas-under-curve (AUCs) with EMD extracted
features, PCA scores and combined features are 0.83, 0.70, and 0.87,
respectively.

\begin{figure}[htbp]
\centering
\includegraphics[width=\linewidth]{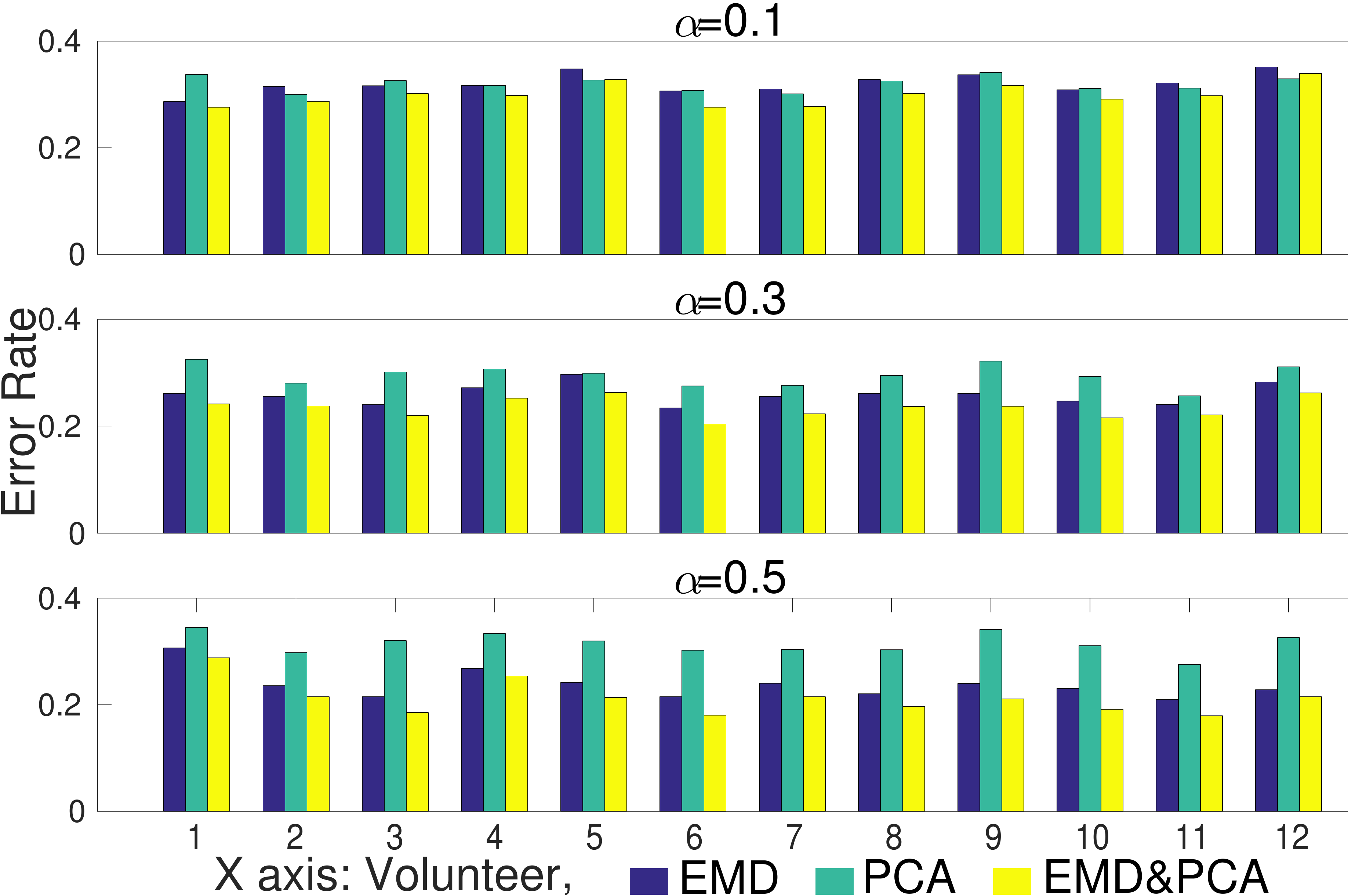}
\caption{The test-set (generalized) detection error for each volunteer, averaged over 50 data
  sets, for three choices of $\alpha$.}
\label{fig:error_each_volunteer}
\end{figure}

We are also interested in examining whether the detection performance is
consistent across volunteers. It is important to check that the
algorithms do not consistently fail to detect tumours for one of the
volunteers.  Figure \ref{fig:error_each_volunteer} shows the detection
error, averaged over the 50 constructed datasets, for three choices of
$\alpha$.

\begin{figure}[htbp]
\includegraphics[width=\linewidth]{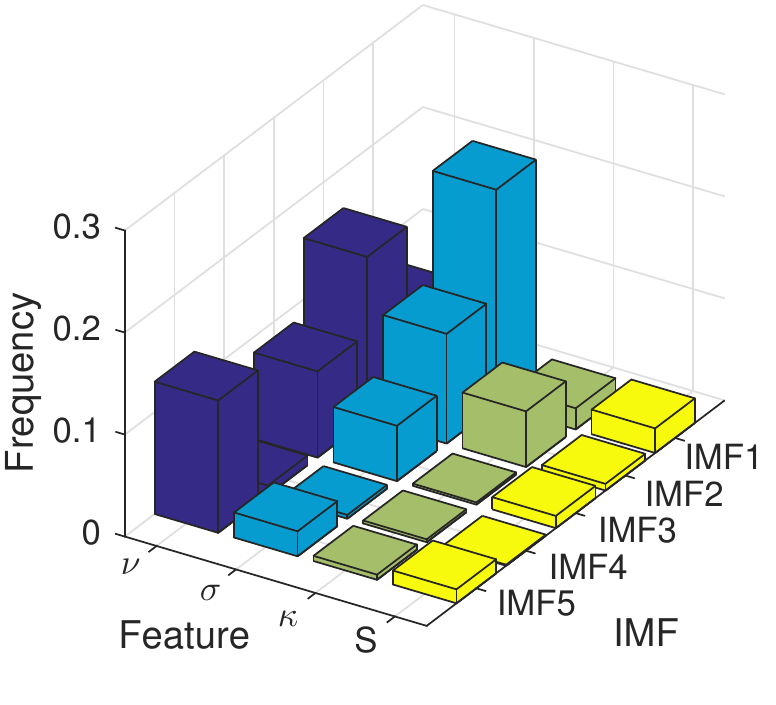}
\centering
\caption{Distribution of the features that are selected by the ensemble
  classifier for the case when the EMD features are separated into
  individual scalar features.}
\label{fig:feature_distribution}
\end{figure}

\begin{figure}[htbp]
\includegraphics[width=\linewidth]{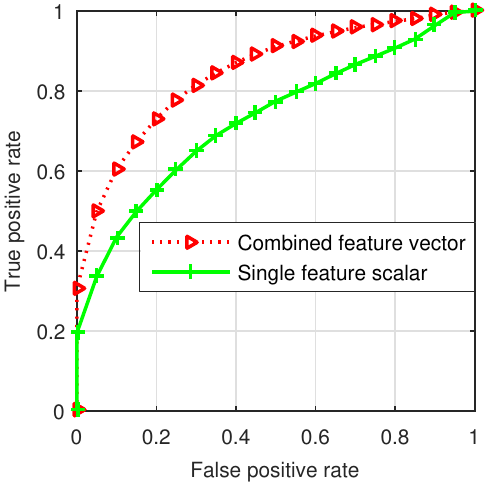}
\centering
\caption{The ROC curves, averaged over 20 datasets with~$\Gamma=1$, for the cases of (i) a combined EMD and PCA feature vector
  (50 elements) and (ii) 20 individual EMD scalar features.}
\label{fig:ROC_singlefeature}
\end{figure}

To investigate which EMD features are most useful,
we separate the $1\times20$ EMD-extracted feature vector for each antenna pair
into 20 scalar features.Thus, the ensemble selection algorithm will be able
to select those EMD features that are most useful in reducing the
training errors. The experiment is carried out on eight randomly selected data sets.
Figure~\ref{fig:feature_distribution} displays the frequency of different features
selected by the ensemble selection classifier.
We observe that the mean absolute value and standard deviation from 5 IMFs are most likely to be selected.
We compare the results from combined features from PCA and EMD with the results from scalar
features. The ROC curves, averaged over the 20 data sets, are shown in Figure \ref{fig:ROC_singlefeature}.

\begin{figure}[htbp]
\centering
\includegraphics[width=\linewidth]{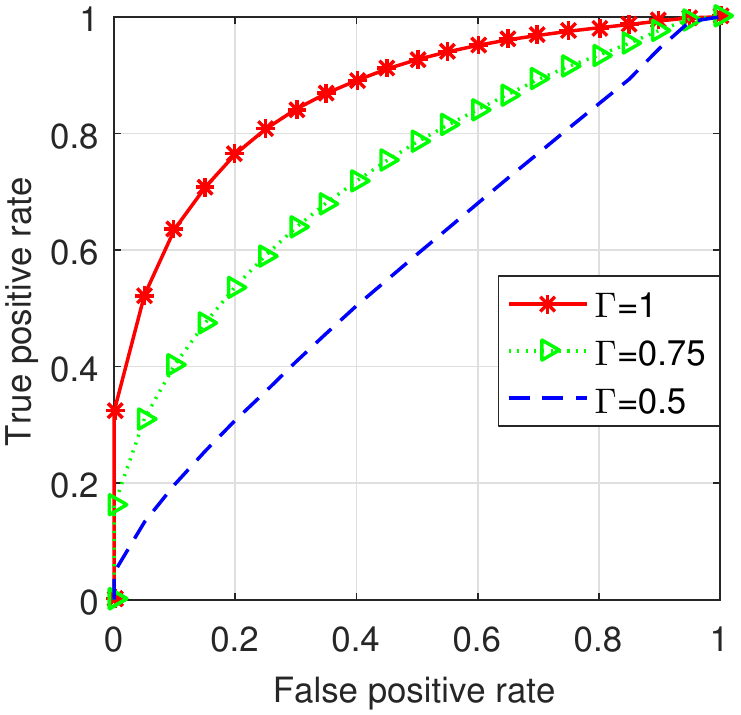}
\caption{The ROC curves, averaged over 20 data sets, for the case of
  combined EMD and PCA features, for three different values of $\Gamma$.}
\label{fig:different_Gamma}
\end{figure}

The final investigation we conducted is to assess the performance of
the combined features (EMD and PCA) on data sets with lower
signal-to-noise ratio (SNR). The lower SNR signals can be constructed
by reducing the value of $\Gamma$ in
Equation~\eqref{equation:tumour_response}.  We generated 20 data sets
with $\Gamma=0.75$ and 20 data sets with $\Gamma=0.5$, respectively.
The ROC curves, averaged over the 20 data sets, are shown in
Figure~\ref{fig:different_Gamma}.

\section{Discussion}
\label{sec:discussion}

We observe from results reported in Section~\ref{sec:experiment results}
that the use of EMD-extracted features results
in significantly better detection performance compared to features
based on PCA. A slight improvement over EMD features can be achieved by using the
combined EMD and PCA features.
As illustrated in Figure~\ref{fig:ROC_singlefeature},
better detection performance is observed when all EMD-extracted features from one antenna pair
are combined into a single feature vector. 

The average detection error is similar for all volunteers,
irrespective of the features employed, as shown in Figure \ref{fig:error_each_volunteer}. We have also
observed (results not shown) that the detection algorithms have consistent performance
with respect to tumour locations, which are randomly generated in the
upper outer quadrant of the breast hemispherical model.

It is observed in Figure~\ref{fig:different_Gamma} that the reduced
signal-to-noise ratio leads to significantly poorer detection
performance, but the performance degradation is graceful. When $\Gamma
= 0.5$, the detection performance is only slightly better than a
random guess.  Although experiments with phantoms suggest that a
choice of $\Gamma = 1$ is reasonable, the appropriate value is
dictated by the contrast between tumorous and healthy tissue, and
there is still considerable uncertainty about the extent of this
contrast. The performance deterioration motivates the development of
feature extraction methods and detection algorithms that can better
tackle these lower signal-to-noise scenarios.

\section{Conclusion}
\label{sec:conclusion}
In this paper, we proposed and explored the use of an EMD-based
feature extraction method for microwave breast tumour detection. We
were motivated to explore the use of EMD features because of their
potential robustness to the system jitter common in microwave breast
cancer scans.  We evaluated detection performance using EMD-based
features and the commonly-used PCA features using a clinical trial data
collected over an eight-month span combined with a numerical tumour
response construction method.
We observe that the use of features based on the EMD leads to
significantly improved detection performance compared to PCA-based
features. A marginal additional improvement can be achieved by using a
combination of EMD and PCA features.

\begin{acknowledgements}
Hongchao Song is sponsored by the China Scholarship Council and the Postgraduate Innovation Fund of SICE, BUPT, 2015. 
\end{acknowledgements}

\bibliographystyle{spmpsci}      
\bibliography{JMBE_v3_Mark_MBEC}   

\begin{authorbiography}{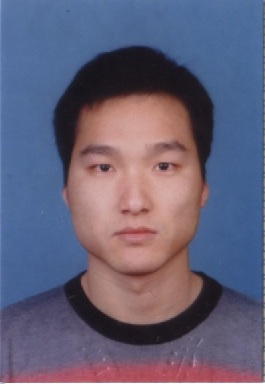}
{Hongchao Song}~is currently a phD candidate in Beijing University of Posts and Telecommunications. From 2015 to 2016, He was a visiting Ph.D. student with the McGill University. His research interests include microwave-based breast cancer detection and digital image processing.
\end{authorbiography}

\begin{authorbiography}{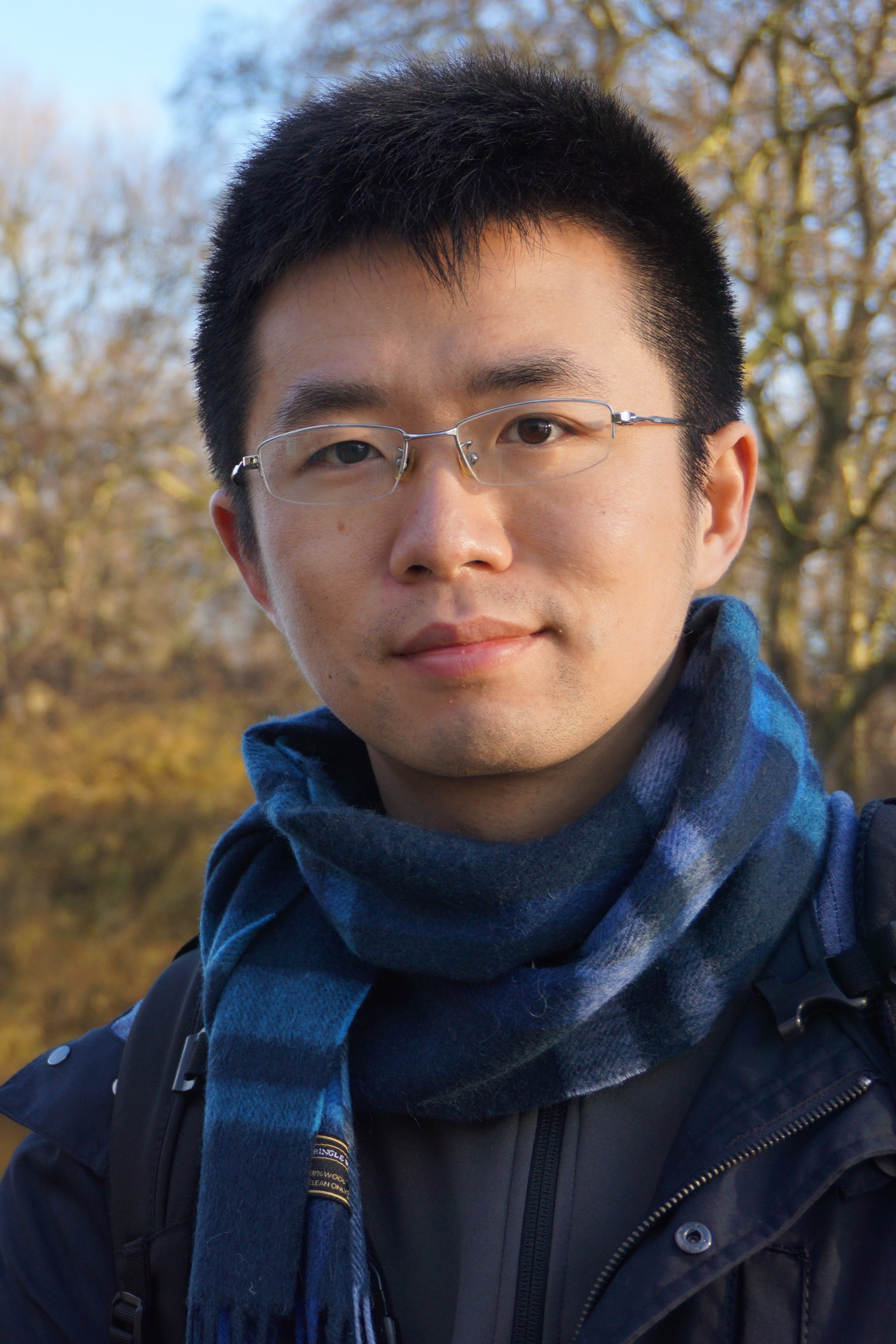}
{Yunpeng Li}~joined McGill University to pursue a Ph.D. degree in electrical engineering, from September 2012. His current research is on microwave-based breast cancer detection and particle filtering.
\end{authorbiography}

\begin{authorbiography}{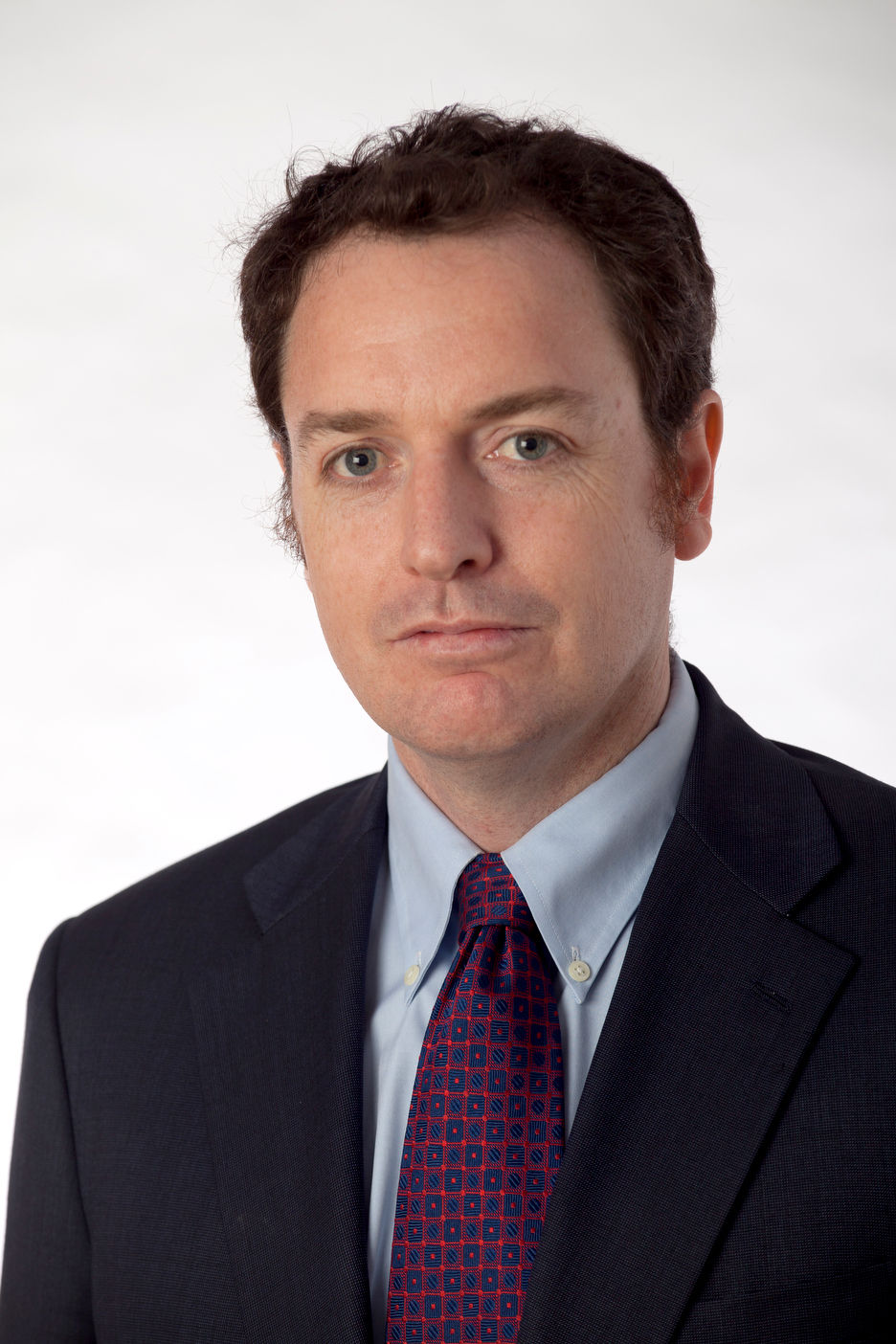}{Mark Coates}~is an Associate Professor at McGill University, Canada. He received a B.E. from the University of Adelaide, Australia (1995), and a Ph.D. from the University of Cambridge, U.K., in 1999.
\end{authorbiography}

\begin{authorbiography}{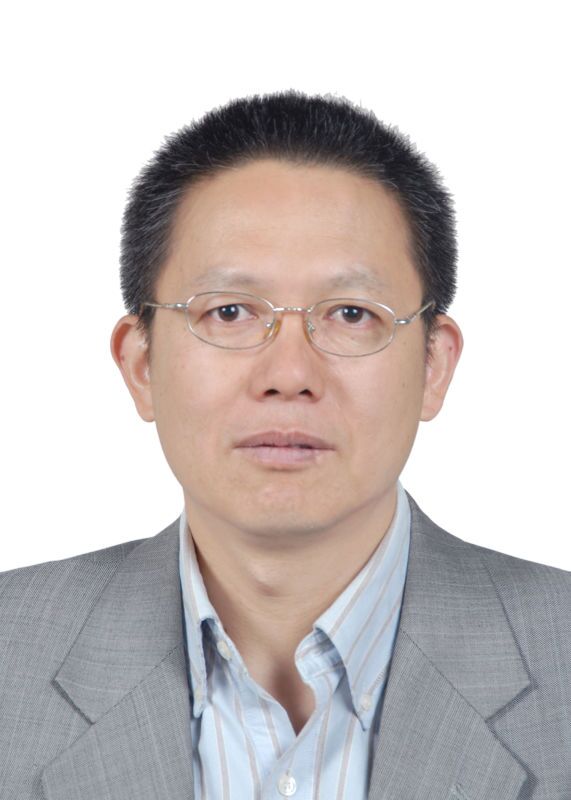}
{Aidong Men}~is a Professor at Beijing University of Posts and Telecommunications, China, where he received his Ph.D in 1994.
\end{authorbiography}

\end{document}